# Tighter Linear Program Relaxations for High Order Graphical Models

**Elad Mezuman**[*]
Hebrew University
Jerusalem, Israel

**Daniel Tarlow**[*]
Microsoft Research
Cambridge, UK

**Amir Globerson**
Hebrew University
Jerusalem, Israel

**Yair Weiss**
Hebrew University
Jerusalem, Israel

## Abstract

Graphical models with High Order Potentials (HOPs) have received considerable interest in recent years. While there are a variety of approaches to inference in these models, nearly all of them amount to solving a linear program (LP) relaxation with *unary consistency* constraints between the HOP and the individual variables. In many cases, the resulting relaxations are loose, and in these cases the results of inference can be poor. It is thus desirable to look for more accurate ways of performing inference. In this work, we study the LP relaxations that result from enforcing additional consistency constraints between the HOP and the rest of the model. We address theoretical questions about the strength of the resulting relaxations compared to the relaxations that arise in standard approaches, and we develop practical and efficient message passing algorithms for optimizing the LPs. Empirically, we show that the LPs with additional consistency constraints lead to more accurate inference on some challenging problems that include a combination of low order and high order terms.

## 1 Introduction

Graphical models are an excellent tool for expressing models that arise in a wide variety of domains including computational biology, natural language processing, and computer vision. A long-standing research challenge is to expand the range of problems that can be expressed with graphical models such that learning and inference can be performed efficiently. Recently, there has been a resurgence of interest in

[*]Equal contribution

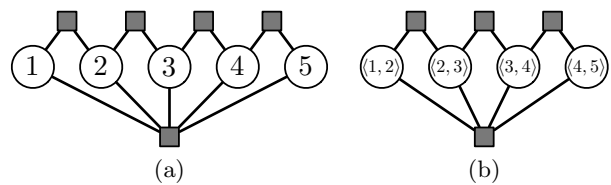

Figure 1.1: We show that LP-based message passing in graphical models with high order potentials (HOPs) is likely to be poor when the HOP communicates via a single-variable interface as in (a). Communicating with a subset of edges of the same model leads to much better results. For example, in (b), it is provably exact.

*high order potentials* (HOPs), which generally refer to modeling components that have tractable structure that is not revealed by looking at the graphical structure of the model. These approaches are rooted in earlier work on graphical models like Pearl's polytree algorithm (Pearl, 1988), noisy-OR interactions (Heckerman, 1989), and context specific independencies (Boutilier et al., 1996). Their recent popularity has been fueled by the availability of efficient routines for using HOPs within modern linear program (LP)-based message passing algorithms such as dual decomposition (Komodakis et al., 2007), MPLP (Globerson and Jaakkola, 2007), and convex belief propagation (Weiss et al., 2007). Recent works such as (Tarlow et al., 2010) attempt to categorize general classes of HOP and give efficient algorithms for using them within message passing algorithms. There are many specific applications, such as using HOPs in an image segmentation task to jointly optimize over the appearance model and segmentation (Vicente et al., 2009).

Despite the success that has been achieved with message passing algorithms and HOPs, nearly all of the approaches are equivalent to a particular LP relaxation (Koller and Friedman, 2009). In addition, several other approaches to dealing with HOPs, such as reducing them to low order models, equate to the same relaxation. We expand on these equivalences in Sec. 4.

The main goal of this paper is to show that the LP relaxation resulting from the standard approach is weak, and to propose an alternative that maintains many of the same desirable computational efficiencies while leading to more accurate inference.

The paper proceeds as follows. We begin by establishing the ubiquity of the relaxation that we term the *unary consistency LP*, showing that many approaches to dealing with HOPs are equivalent to this relaxation. Having established this, we go on to show that the unary consistency relaxation is quite weak. We provide several examples and some analysis to help understand when this is the case and why it fails. Next, we introduce a family of LPs that provide tighter relaxations than the standard relaxation, but (as we show) still admit efficient algorithms for optimizing them. We provide theoretical analysis of these LPs, showing when they are provably tight.

We then turn to practical concerns and show that our tighter LPs can often be efficiently solved using standard message-passing algorithms: the main difference is that the nodes corresponding to the higher order potential now receive and send messages that are functions of pairs of variables, not just singletons. This makes the message-passing more involved but we identify special cases of HOPs for which the message computation is still tractable. We also show how to use the messages to compute tighter bounds on the MAP and how to choose which pairs of variables should be added in a way that is guaranteed to tighten the bound. We illustrate the performance of our method on both synthetic models and real image segmentation problems.

## 2 Motivating Examples

To begin, we will establish notation, then consider two concrete examples that illustrate the looseness of the standard LP relaxation for dealing with HOPs.

**Notation and Preliminaries** We let an energy function over $n$ discrete variables $\mathbf{x} = \{x_1, ..., x_n\}$ be defined as follows. Given a graph $G = (\mathcal{V}, \mathcal{E})$ with $n$ vertices, there are potentials $\theta_i(x_i), \theta_{ij}(x_i, x_j)$ for each vertex and each edge in the graph, respectively, and one HOP over all the variables, $\theta_\alpha(\mathbf{x})$. We wish to find the minimum energy configuration (alternatively the *maximum a posteriori* (MAP) assignment):

$$\mathbf{x}^* = \arg\min_{\mathbf{x}} \sum_{i \in \mathcal{V}} \theta_i(x_i) + \sum_{ij \in \mathcal{E}} \theta_{ij}(x_i, x_j) + \theta_\alpha(\mathbf{x}). \quad (2.1)$$

This is an integer program that is NP hard to solve in general. The LP relaxation approach works in two stages. First, the integer program is converted into a linear program with the same solution, but which requires exponentially many constraints to express the domain. Then the linear program is relaxed by outer bounding the domain to yield the $\mathbf{LP}_\emptyset$ below. For more background on LP relaxations, we recommend Wainwright and Jordan (2008), Koller and Friedman (2009) and Sontag et al. (2010).

We now review the standard LP relaxation approach to approximating Eq. 2.1. The relaxation maintains three types of distributions: over single variables, over pairs of variables, and over all of $x$. All three are constrained to agree on their singleton marginals. The singleton, pairwise and HOP terms are then replaced by their expectation according to the corresponding distributions. We call this approach the *Unary Consistency LP* and denote it by $\mathbf{LP}_\emptyset$. The resulting optimization problem is

**LP$_\emptyset$ (Unary Consistency LP)**

$$\min_q \sum_{i \in \mathcal{V}} \mathbb{E}_{q_i}[\theta(x_i)] + \sum_{ij \in \mathcal{E}} \mathbb{E}_{q_{ij}}[\theta_{ij}(x_i, x_j)] + \mathbb{E}_{q_\alpha}[\theta_\alpha(\mathbf{x})]$$

$$\text{s.t.} \sum_{x_i} q_{ij}(x_i, x_j) = q_j(x_j) \, , \, \sum_{x_j} q_{ij}(x_i, x_j) = q_i(x_i)$$

$$\sum_{\mathbf{x}:\mathbf{x}(i)=x_i} q_\alpha(\mathbf{x}) = q_i(x_i) \qquad \forall i \in \mathcal{V}, \quad (2.2)$$

where we omit for space (as we will throughout) the additional constraints that $q$ variables are non-negative and sum to 1, and some of the quantifications (e.g., $ij \in \mathcal{E}$ and the values of $x_i$ in the last constraint). As we show in Sec. 4, this relaxation is commonly used because it can be solved efficiently for several families of HOPs.

**Introductory Examples** We start with two simple instances of MAP problems with tractable HOP for which the standard approach fails. In both cases the factor graph (Fig. 1.1a) consists of a simple chain and a cardinality-based potential which is amenable to the techniques described in (Tarlow et al., 2010) for solving $\mathbf{LP}_\emptyset$. Unfortunately, the relaxation is loose, which is manifested in a fractional solution that has better objective than the original integer program solution. This means the relaxation is inaccurate, and, more importantly, it prevents us from finding a good solution to the original problem.

In both examples $x_i \in \{0, 1\}$ and the model is a chain with an even number of nodes and attractive pairwise potentials of the form $\theta_{i,i+1} = \begin{pmatrix} 0 & c \\ c & 0 \end{pmatrix}$ (where $c > 0$).

1. Finding the 2nd best assignment (Fromer and Globerson, 2009) using an exclusion factor. If we add a local potential that favors the binary variables being off: $\theta_i = \begin{pmatrix} 0 & \epsilon \end{pmatrix}^\top$ with $\epsilon > 0$, then clearly the best assignment is $\mathbf{x} = \vec{0}$. We add an exclusion potential to exclude this assignment: $\theta_\alpha(\mathbf{x}) = \infty$ when $\mathbf{x} = \vec{0}$ and 0 otherwise. and now the MAP of the problem with the high-order potential is the second best assignment in the original problem. Assuming $c \gg \epsilon$, the $2^{nd}$ best is $\mathbf{x} = \vec{1}$ with an energy of $n\epsilon$. However, the $\mathbf{LP}_\emptyset$ solution has a value of $\epsilon$ and corresponds to the following fractional optimum: $q_i = \begin{pmatrix} \frac{n-1}{n} & \frac{1}{n} \end{pmatrix}^\top$, $q_{i,i+1} = \begin{pmatrix} \frac{n-1}{n} & 0 \\ 0 & \frac{1}{n} \end{pmatrix}$, and $q_\alpha(\mathbf{x}) = \frac{1}{n}$ for all the assignments in which exactly one variable is on.

2. Partitioning graphs using average-cut (Mezuman and Weiss, 2012). Here the HOP prefers assignments with similar number of off and on variables: $\theta_\alpha(\mathbf{x}) = -\lambda \cdot |\mathbf{x}| \cdot (n - |\mathbf{x}|)$. The optimal integral solution is to break the chain in the middle with value of $c - \lambda(\frac{n}{2})^2$, but the $\mathbf{LP}_\emptyset$ solution has a value of $-\lambda(\frac{n}{2})^2$ and is again fractional: $q_i = \begin{pmatrix} 0.5 \\ 0.5 \end{pmatrix}$, $q_{i,i+1} = \begin{pmatrix} 0.5 & 0 \\ 0 & 0.5 \end{pmatrix}$, and $q_\alpha(\mathbf{x}) = \frac{1}{M}$ for all $M = \binom{n}{n/2}$ assignments that have $n/2$ ones.

Notice that the LP solution is not only fractional, but is in fact also completely "tied": its solution gives us no hint as to the true MAP.

## 3 Related work

**Applications of High Order Potentials** HOPs can be used to incorporate nonlocal structure into a model. In recent years, there have been many works that incorporate these types of interactions. They are particularly useful for modelling highly structured global interactions like those that arise in models for parsing sentences (Smith and Eisner, 2008; Koo et al., 2010; Martins et al., 2010), in models for image segmentation to enforce connectivity constraints (Nowozin and Lampert, 2009) or higher order smoothness (Kohli et al., 2007; Gould, 2011), and in models of textures to encourage soft pattern matching (Rother et al., 2009). They arise when "collapsing" certain models, like in (Vicente et al., 2009), where optimizing out an image segmentation appearance model leads to an energy function over segmentations that has high order terms. They have also been used to solve balanced graph partitioning problems (Mezuman and Weiss, 2012) and to enforce constraints over latent variable activations in e.g., Restricted Boltzmann Machines (Swersky et al., 2012).

**Tighter Linear Program Relaxations** The canonical works on tightening LP relaxations using message-passing come from Sontag et al. (2008); Werner (2008) and Komodakis and Paragios (2008), and were followed up in several works, such as Batra et al. (2011); Sontag et al. (2012). As discussed in Sontag (2010), at their core, these approaches can be viewed as searching in similar ways for additional consistency constraints to enforce such that adding them to the LP leads to a tighter relaxation. While the general approach is applicable in the LP relaxations with HOPs that we consider here, there are computational challenges that must be addressed in order to do this search efficiently. We develop the needed methods in this work. We note that while Werner (2008) discusses HOPs in the context of the max-sum diffusion algorithm and the class of LPs that we study here can be expressed within the framework presented there, the final suggestion for working with HOPs is to use the unary consistency LP, and a proof is provided that if the model is submodular and the HOP is submodular, then the LP is tight.

Fromer and Globerson (2009) deal with the case of excluding a single joint assignment. We will show that the baseline they consider, Santos, is equivalent to unary consistency, and that their method (which leads to tight relaxations on trees) is a special case of our method. Thus, we can get an equally tight relaxation by using the approach proposed in this paper. Another special case of note where tighter HOP relaxations have been discussed is Komodakis and Paragios (2009). There, they employ a merging strategy for dealing with several pattern-based HOPs, in that they show that patterns along rows and columns of a grid can be combined into a single HOP where messages can still be computed efficiently. However, after this merging, only unary consistency is enforced. For the purposes of this paper, we assume throughout that if any tractability-preserving merging of HOPs is possible, it has already been done.

## 4 Unary Consistency Linear Programs

Many existing methods for inference with HOPs are equivalent to $\mathbf{LP}_\emptyset$. We highlight some of these below.

**Message Passing with the standard Factor Graph** Perhaps the most common method for solving MAP inference in graphical models with HOPs is to build the factor graph and pass messages between variables and factors. For certain higher-order potentials, the messages between the variable nodes and the HOP node can be calculated efficiently (Tarlow et al., 2010).

Different works use somewhat different update schemes. One option is to use loopy belief propagation (Yedidia et al., 2005). However, there are typi-cally no performance guarantees in this case (e.g., no convergence results or optimality certificates). A different class of structurally similar algorithms retain the message-passing flavor of BP while also giving an optimality certificate (Globerson and Jaakkola, 2007; Werner, 2007; Kolmogorov, 2006; Komodakis et al., 2010; Sontag et al., 2010; Weiss et al., 2007). These can all be shown to be solving $\mathbf{LP}_\emptyset$.

**Simplification with Auxiliary Variables** An alternative strategy for dealing with certain HOPs is to create auxiliary variables in such a way as to reduce the problem to a pairwise problem, and then solve the pairwise problem using the standard pairwise LP relaxation. Here, we study the strength of the LP relaxations that result from this strategy. For example, Kohli et al. (2009), Gould (2011), and Rother et al. (2009) all follow this or a closely related approach.

The approach proceeds as follows. Start with a HOP and some unary and pairwise potentials:

$$E(\mathbf{x}) = \sum_i \theta_i(x_i) + \sum_{ij} \theta_{ij}(x_i, x_j) + \theta_\alpha(\mathbf{x}). \quad (4.1)$$

Next, introduce an auxiliary variable $z$ such that minimizing it out leaves the energy over $\mathbf{x}$ unchanged. Namely, $\min_z \theta_{z,\alpha}(z, \mathbf{x}) = \theta_\alpha(\mathbf{x})$. We then have:

$$E(\mathbf{x}) = \min_z \sum_i \theta_i(x_i) + \sum_{ij} \theta_{ij}(x_i, x_j) + \theta_{z,\alpha}(z, \mathbf{x}).$$

Finally, it often holds that given $z$, the HOP becomes fully factorized, i.e., $\min_z \theta_{z,\alpha}(z, \mathbf{x}) = \min_z \sum_i \theta_{zi}(z, x_i)$, so that $E(\mathbf{x})$ is given by:

$$\min_z \sum_i \theta_i(x_i) + \sum_{ij} \theta_{ij}(x_i, x_j) + \sum_i \theta_{zi}(z, x_i). \quad (4.2)$$

At this point, the minimization over $x$ can be done jointly with the minimization over $z$ using the following LP relaxation, which we call $\mathbf{LP_{red}}$:

**$\mathbf{LP_{red}}$**

$$\min_q \sum_{i\in\mathcal{V}}(\mathbb{E}_{q_i}[\theta(x_i)] + \mathbb{E}_{q_{zi}}[\theta_{zi}(z,x_i)]) + \sum_{ij\in\mathcal{E}}\mathbb{E}_{q_{ij}}[\theta_{ij}(x_i,x_j)]$$

$$\sum_{x_j} q_{ij}(x_i, x_j) = q_i(x_i) \quad \sum_{x_i} q_{ij}(x_i, x_j) = q_j(x_j) \quad (4.3)$$

$$\sum_z q_{zi}(z, x_i) = q_i(x_i) \quad \sum_{x_i} q_{zi}(z, x_i) = q_z(z) \quad (4.4)$$

This relaxation seems quite different from $\mathbf{LP}_\emptyset$. However, we have the surprising result that they are in fact equivalent (the proof is in the appendix).

**Proposition 1.** *The relaxations $\mathbf{LP}_\emptyset$ and $\mathbf{LP_{red}}$ are equivalent. Namely, they have the same objective value, and there is a mapping between their optima.*

In the case that more than one auxiliary variable is created by the pairwise transformation, the unary consistency LP will be at least as tight as the reduced LP, but they are no longer equal in general. A corollary of this analysis is that if the pairwise transformation introduces only submodular pairwise terms (and the pairwise part of the model is submodular), then the unary consistency LP is tight. This is closely related to (but less general than) the result proved in (Werner, 2008).

**Exclusion Potentials and the Santos Inequality** A special case of HOP model that has received significant attention is where a model is modified so as to exclude a single joint assignment $\mathbf{x}^*$ (e.g., the first introductory example). In this context, several LP relaxations have been proposed (e.g., Fromer and Globerson, 2009). It is thus interesting to ask which of these is equivalent to $\mathbf{LP}_\emptyset$. It turns out that $\mathbf{LP}_\emptyset$ corresponds to an LP with no $q_\alpha(\mathbf{x})$ variables, but rather a single constraint (in addition to the pairwise consistency constraints): $\sum_i q_i(x_i^*) \leq n - 1$, which was first suggested in Santos Jr (1991). Intuitively, it states that at most $n - 1$ of the variables can agree with the assignment $\mathbf{x}^*$. The proof of the equivalence to $\mathbf{LP}_\emptyset$ is straightforward and follows from the characterization of the assignment excluding polytope for an empty graph, and its relation to the Santos inequality (see Fromer and Globerson, 2009).

## 5 Tighter Linear Programs

In this section, we introduce the family of tighter LP relaxations that are the focus of this work, and we study their theoretical properties. We begin by defining a family of LPs that are tighter relaxations than $\mathbf{LP}_\emptyset$, and then we will prove a tightness result.

Let $S \subseteq \mathcal{E}$ be a subset of the edges in $G$, and define $\mathcal{V}(S)$ to be the set of variables that appear in at least one edge in $S$. We can then define an LP that enforces consistency between the HOP and the edges in $S$, while maintaining unary consistency with variables in $\mathcal{V} - \mathcal{V}(S)$:

**$\mathbf{LP}_S$ (Partial Edge Consistency LP)**

$$\min_q \sum_{i\in\mathcal{V}} \mathbb{E}_{q_i}[\theta(x_i)] + \sum_{ij\in\mathcal{E}} \mathbb{E}_{q_{ij}}[\theta_{ij}(x_i, x_j)] + \mathbb{E}_{q_\alpha}[\theta_\alpha(\mathbf{x})]$$

$$\text{s.t.} \sum_{x_i} q_{ij}(x_i, x_j) = q_j(x_j) \, , \, \sum_{x_j} q_{ij}(x_i, x_j) = q_i(x_i)$$

$$\sum_{\mathbf{x}: \mathbf{x}(i) = x_i} q_\alpha(\mathbf{x}) = q_i(x_i) \quad \forall i \in \mathcal{V} - \mathcal{V}(S)$$

$$\sum_{\mathbf{x}: \mathbf{x}(i) = x_i, \mathbf{x}(j) = x_j} q_\alpha(\mathbf{x}) = q_{ij}(x_i, x_j) \quad \forall ij \in S$$

At one extreme, when $S = \emptyset$, $\mathbf{LP}_S$ is equal to $\mathbf{LP}_\emptyset$. At the other extreme, consistency is enforced between the HOP and all edges, yielding the following simplified LP, which will be of special interest:

**$\mathbf{LP}_\mathcal{E}$ (Full Edge Consistency LP)**

$$\min_q \sum_{i \in \mathcal{V}} \mathbb{E}_{q_i}[\theta(x_i)] + \sum_{ij \in \mathcal{E}} \mathbb{E}_{q_{ij}}[\theta_{ij}(x_i, x_j)] + \mathbb{E}_{q_\alpha}[\theta_\alpha(\mathbf{x})]$$

$$\text{s.t. } \sum_{x_i} q_{ij}(x_i, x_j) = q_j(x_j) \,,\, \sum_{x_j} q_{ij}(x_i, x_j) = q_i(x_i)$$

$$\sum_{\mathbf{x}:\mathbf{x}(i)=x_i,\mathbf{x}(j)=x_j} q_\alpha(\mathbf{x}) = q_{ij}(x_i, x_j) \quad \forall ij \in \mathcal{E}$$

### 5.1 Strength of $\mathbf{LP}_\mathcal{E}$

We begin with the simple observation that $\mathbf{LP}_\mathcal{E}$ is always tight.

**Proposition 2.** $\mathbf{LP}_\mathcal{E}$ *is tight.*

*Proof.* Since with all the pairwise constraints the expectation of $\theta_i$ and $\theta_{ij}$ under $q_\alpha$ is the same as under $q_i$ and $q_{ij}$, respectively, $\mathbf{LP}_\mathcal{E}$ is equivalent to

$$\min_q \mathbb{E}_{q_\alpha}[\theta_\alpha(\mathbf{x}) + \theta(x_i) + \theta_{ij}(x_i, x_j)]. \quad (5.1)$$

This LP has an integer solution (i.e., it is tight) because it is always better to put all the $q_\alpha$ mass on the best assignment than to divide it. $\square$

Thus, the space of LP relaxations that can be constructed as $\mathbf{LP}_S$ for some choice of edge set $S$ range from the standard but weak $\mathbf{LP}_\emptyset$ to the tight $\mathbf{LP}_\mathcal{E}$. This justifies our focus on this family of $\mathbf{LP}_S$.

## 6 Optimization with Message Passing

As we reviewed earlier, $\mathbf{LP}_\emptyset$ can be solved by a variety of message-passing algorithms operating on the standard factor graph, where factor nodes communicate with individual variable nodes. Similarly, it is easy to show that the same algorithms can solve $\mathbf{LP}_S$ when they are applied on a modified factor graph where the node corresponding to the HOP communicates with *pairs of nodes* which correspond to edges in $S$ (see Fig. 1.1b). The key question is the complexity of calculating the messages to and from the HOP. We now identify cases where these messages can be computed efficiently.

We begin by recalling the dual of $\mathbf{LP}_S$. The dual variables are $\delta_{ij}(x_i, x_j)$ for $ij \in S$ (which we interpret as messages between the factor $\alpha$ and the edges in $S$) and $\delta_i(x_i)$ for $i \in \mathcal{V}$ (messages between the factor $\alpha$ and singletons). The dual problem is to maximize $B(\delta)$, a lower bound on the MAP:

$$B(\delta) = \sum_i \min_{x_i} \tilde{\theta}_i^\delta(x_i) + \sum_{ij} \min_{x_i, x_j} \tilde{\theta}_{ij}^\delta(x_i, x_j) + \min_x \tilde{\theta}_\alpha^\delta(\mathbf{x}),$$
(6.1)

where $\tilde{\theta}^\delta$ is a reparameterization of the original energy function:

$$\tilde{\theta}_\alpha^\delta(\mathbf{x}) = \theta_\alpha(\mathbf{x}) - \sum_{ij \in S} \delta_{ij}(x_i, x_j) - \sum_{i \in \mathcal{V} - \mathcal{V}(S)} \delta_i(x_i).$$

Expressions for $\tilde{\theta}_i^\delta(x_i), \tilde{\theta}_{ij}^\delta(x_i, x_j)$ are similar (Sontag et al., 2010). Most message passing approaches for solving this problem iteratively update $\delta$ to increase the bound. All these message update schemes require solving $\min_x \tilde{\theta}_\alpha^\delta(\mathbf{x})$ for arbitrary values of $\delta$, or calculating its min-marginals (see Sontag et al., 2010, for a thorough review of such approaches). For general $\theta_\alpha(\mathbf{x})$ this is of course difficult. Below we highlight some cases where it is tractable, and therefore $\mathbf{LP}_S$ can be solved efficiently with message passing.

- Low tree-width $S$ graphs with cardinality-based potentials. When $\theta_\alpha(\mathbf{x})$ is a cardinality potential (i.e., $\theta_\alpha(\mathbf{x}) = f(\sum_i x_i)$, where $f(\cdot)$ is some arbitrary function) and $S$ is a tree-structured subset of edges, then this is closely related to one of the problems considered by (Tarlow et al., 2012). There, it was shown that messages to and from the HOP can be calculated by performing exact inference on an augmented tree graph with complexity that is at most $O(n^2)$. This result is easily extended to the case where $S$ forms a low tree-width graph and the messages can be computed exactly in a time that is exponential in the tree-width of $S$.

- Low tree-width $S$ graphs with Pattern HOPs. Another HOP that has received interest are the pattern potentials of (Rother et al., 2009). Here, the potentials are of the form $\theta_\alpha(\mathbf{x}) = \min_{k \in \{1,...,K\}} \sum_i w_i^{(k)} x_i$ where each real-valued vector $w^{(k)}$ can be thought of as encoding a pattern that is desirable to match. This potential is actually quite simple to work with, by noting that $\min_x \tilde{\theta}_\alpha^\delta(\mathbf{x})$ is equivalent to:

$$\min_{k,x} \left[ \sum_i w_i^{(k)} x_i - \sum_{ij \in S} \delta_{ij}(x_i, x_j) - \sum_{i \in \mathcal{V} - \mathcal{V}(S)} \delta_i(x_i) \right].$$

From here, it is clear that the argmin or the min-marginals can be computed by constructing a junction tree over $S$, then solving $K$ different problems where for problem $k$, the unary potentials have been modified by $w^{(k)}$, then taking the elementwise minima.

A simple corollary of the above discussion is that whenever the graphical model has low tree-width and the HOP is either cardinality-based or a pattern HOP then $\mathbf{LP}_\mathcal{E}$ can be efficiently computed. In particular, for the two motivating examples discussed in the introduction, running message-passing on the factor graph shown in Fig. 1.1b can be performed efficiently and is guaranteed to provide the MAP.

## 7  Choosing a Tractable Edge Set

When the graph has high tree-width, we cannot efficiently solve $\mathbf{LP}_\mathcal{E}$, and a natural question is how to choose a subset of edges $S$ such that $\mathbf{LP}_S$ is as tight as possible, but can still be solved in practice. For the HOPs we consider, this will be the case as long as $S$ has low tree-width (see Sec. 6).

As mentioned earlier, this problem has been studied generally in several works, including Sontag et al. (2008), Werner (2008) and Komodakis and Paragios (2009). When trying to adapt these approaches to models with HOPs that are based on $\mathbf{LP}_S$, the general methodology stays the same, but as with the message updates, computational challenges arise. Specifically, the above methods are all based around finding additional consistency constraints to add that are guaranteed to improve $B(\delta)$. In our context, the motivation is the following Lemma:

**Lemma 1.** *Suppose we have solved the dual of $\mathbf{LP}_S$ with some set of edges $S$. If there exists an edge $(i,j) \notin S$ for which there is no overlap between the minimizing assignments of $\tilde{\theta}_{ij}$ and the minimizing assignments of $\tilde{\theta}_\alpha$, then defining $T = \{S \cup (i,j)\}$ we have $\mathbf{LP}_S < \mathbf{LP}_T$ (i.e. adding that edge to $S$ will lead to a strictly tighter LP relaxation).*

The proof follows from using the reparameterization given by the dual variables of $\mathbf{LP}_S$ to construct a valid reparameterization for the dual of $\mathbf{LP}_T$ and the dual value will be strictly higher. Existing methods are often based on similar reasoning (e.g., this is essentially the same result that appears in Werner (2008) in the discussion of cutting planes).

**Sequentially Adding Edges using Weak Cycle Agreement (WCA)**  Recall that our goal is to find a low tree-width $S$ such that $\mathbf{LP}_S$ is as tight as possible. Motivated by Lem. 1 we use the following procedure for approximating such a set. Start with an edge set, $S$, with tree-width one. Then, keep adding edges as long as the tree-width stays small. The edges added are those that satisfy the condition in Lem. 1 and hence result in strict increase of the LP objective. In what follows we provide additional details.

To obtain the initial tree we follow a simple heuristic. Calculate weights $w_{ij} = \max(\theta_{ij}) - \min(\theta_{ij})$ for each edge, and find the spanning tree $S$ with the maximum overall weight. The rationale is that edges with close to uniform potentials (i.e., low $w_{ij}$) are more likely to be consistent with the HOP.

Next, we add $K$ edges in each iteration using the following procedure. Run $\mathbf{LP}_S$ to convergence, and find the *set* of $M$ assignments that minimize the HOP term of the reparameterization: $\{\mathbf{x}^m\}_{m=1}^M = \arg\min_{\mathbf{x}} \tilde{\theta}_\alpha(\mathbf{x})$. As long as $M$ is small—as we found it is in practice—this can be done efficiently by following all back-pointers when decoding from the junction tree structures used for computing message updates. Next, for each edge $ij \in \mathcal{E} \setminus S$ whose addition to $S$ does not violate the maximum tree-width, we compute its *weak cycle agreement* (WCA) measure: $\min_{\mathbf{x}^m} \tilde{\theta}_{ij}(x_i^m, x_j^m) - \min_{x_i, x_j} \tilde{\theta}_{ij}(x_i, x_j)$. By Lem. 1, addition of any edge with WCA $> 0$ will give a tighter relaxation. If many edges have WCA $> 0$, we add the one with the greatest WCA value. Before adding the next $K-1$ edges, we move the reparameterized edge potential into the HOP, recompute the argmins over $\tilde{\theta}_\alpha$, then update the WCAs. Notice that the WCA measure relates to the weak tree agreement (WTA) measure from Tarlow et al. (2011). We do not use the WCA to select the starting set of edges because we have found that in many times after $\mathbf{LP}_\emptyset$ converges, the WCA of all the edges is equal to zero, i.e., there is no single edge whose addition to $S$ will tighten the bound. This rarely happens when $S$ is non-empty.

## 8  Experimental Results

We conducted three sets of experiments over graphical models with different kinds of cardinality HOPs. The first extends the first example in Sec. 2 and shows experimentally that $\mathbf{LP}_\emptyset$ does not find the MAP solution in many simple cases. The second compares our edge selection criterion from Sec. 7 with other possible criteria and shows it is superior to them. The last set of experiments was done over images from the Berkeley segmentation dataset (Martin et al., 2001) and shows we can often find the optimal average-cut (NP-hard problem in the general case) by solving $\mathbf{LP}_S$ with a low tree-width set of edges. The first set of experiments was done on a relatively small problem which allowed us to use a commercial LP solver (Mosek). Since the goal of this experiment was to understand $\mathbf{LP}_\emptyset$ we preferred using it and avoid possible difficulties when solving the dual problem. The other two experiments were solved using message passing with convex belief propagation, as described in (Weiss et al., 2007, 2011), applied to $\mathbf{LP}_S$. We built our junction-tree code on top of the UGM package (Schmidt, 2012).

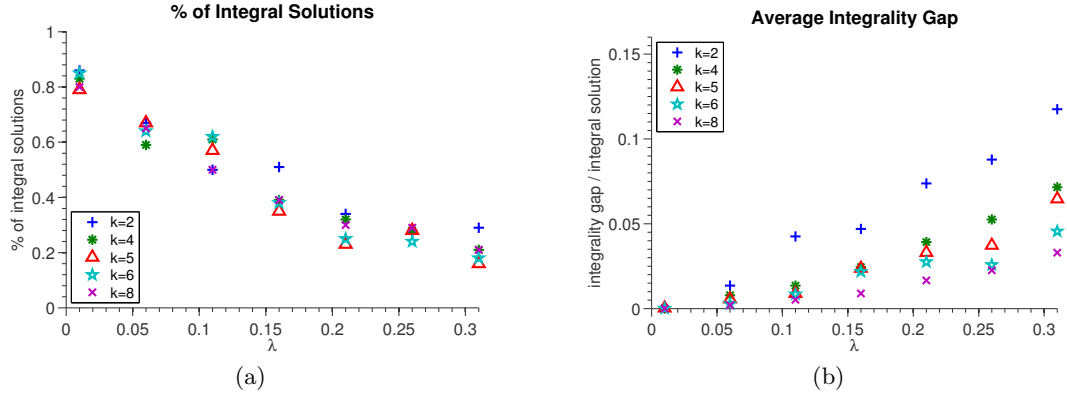

Figure 8.1: $\mathbf{LP}_\emptyset$ Hamming ball exclusion. The 2nd best mode of a tree model with attractive pairwise potentials ($\lambda$) is rarely found when using only unary consistency constraints ($\mathbf{LP}_\emptyset$). We conducted the experiments for several radiuses of hamming balls (k). Using our suggested method it can always be found exactly and efficiently.

**Hamming Ball Exclusion** When finding M-best modes (Batra et al., 2012), it is desirable to have dissimilarity measures that do not factorize into a sum of pixel dissimilarities, e.g., to represent the notion of *dissimilar enough*, where as long as an assignment is at least Hamming distance $k$ away, we are equally happy. In this case, the constraint of Fromer and Globerson (2009) is no longer applicable, so the strategy suggested by Batra et al. (2012), is to add a HOP enforcing this constraint, then solving $\mathbf{LP}_\emptyset$. We know from our theoretical analysis that $\mathbf{LP}_\emptyset$ cannot solve all instances of this problem. Here, we empirically study how bad the relaxation becomes (measured by the integrality gap) as the strength of pairwise potentials is varied. We generated random trees over 10 variables with attractive pairwise potentials $\theta_{ij}(x_i, x_j) = \begin{pmatrix} 0 & \lambda \\ \lambda & 0 \end{pmatrix}$, and unary potentials with random preference to be off ($\theta_i(1) \sim \mathcal{U}[0,1]$); thus the MAP assignment is all zeros. We add an exclusion factor which allows only assignments which are at least $k$ Hamming distance away: $\theta_\alpha(\mathbf{x}) = \infty$ if $|x| < k$ and 0 otherwise. Fig. 8.1 shows the percent of integral solutions and the average integrality gap (out of 100 experiments) for different $k$'s and $\lambda'$s. Notice that while $\mathbf{LP}_\emptyset$ fails in finding the MAP for this problem, the tree-width of $\mathbf{LP}_S$ here is one and thus we can solve it efficiently and exactly.

**Sequentially Adding Edges** We compare the improvements in the dual bound that result from using different criteria for choosing edges to add to $S$ and then solving $\mathbf{LP}_S$. The first criterion is our suggested WCA method, and the second is the potential weight heuristic, both of which are described in Sec. 7. The third criterion is simply adding random edges (RND1 and RND2). We conducted experiments over several 4-connected 7x7 grid, with random pairwise potentials. The HOP is average-cut, $\theta_\alpha(\mathbf{x}) = -\lambda \cdot |\mathbf{x}| \cdot (n - |\mathbf{x}|)$, where $\lambda$ is chosen such that the optimal energy will be zero, and $\theta_1(0) = \infty$ to break the symmetry. We did not limit the tree-width in this experiment. Figure 8.2 shows the improvement in the bound after each edge addition to the starting tree. Clearly the WCA criterion is better than the other two baselines.

**Image Segmentation using super-pixels** Finally, we construct average-cut problems for 40 images from the Berkeley segmentation dataset (Martin et al., 2001) and attempt to solve them using $\mathbf{LP}_S$ and the WCA measure for choosing $S$. We used the procedure described in (Mezuman and Weiss, 2012) to find a setting of $\lambda$ such that solving $\mathbf{LP}_\mathcal{E}$ would verify the optimality of the average-cut. We used SLIC superpixels (Achanta et al., 2012) using the implementation of (Mueller, 2012). We chose SLIC parameters to get approximately 100 equally sized superpixels. The pairwise potentials (affinities between pixels) were computed using intervening contours (Leung and Malik, 1998) (implementation provided by (Cour et al., 2010)).

When choosing $S$, we limited the tree-width to be at most six (the average tree-width of the full graph is 13). We add edges in batches of eight after the previous $\mathbf{LP}_S$ was solved (i.e., the BP converged). Via this procedure we provably found the optimal solution in 34 out of the 40 images. For 2 images the optimum was found when $S$ was of tree-width of 2, for 13 when $S$ was of tree-width 3, and for 10, 5 and 4 images when $S$ was of tree-width of 4, 5 and 6, respectively. Our solution improved the standard spectral solution in 38 out of 40 problems, with an average improvement in the objective of 70%. Fig. 8.4 shows the maximal dual bound achieved during message passing across sets of edges with different tree-widths. To keep the plot clean we show it only for the 25 images for which the $\mathbf{LP}_S$ was tight with tree-width at most four.

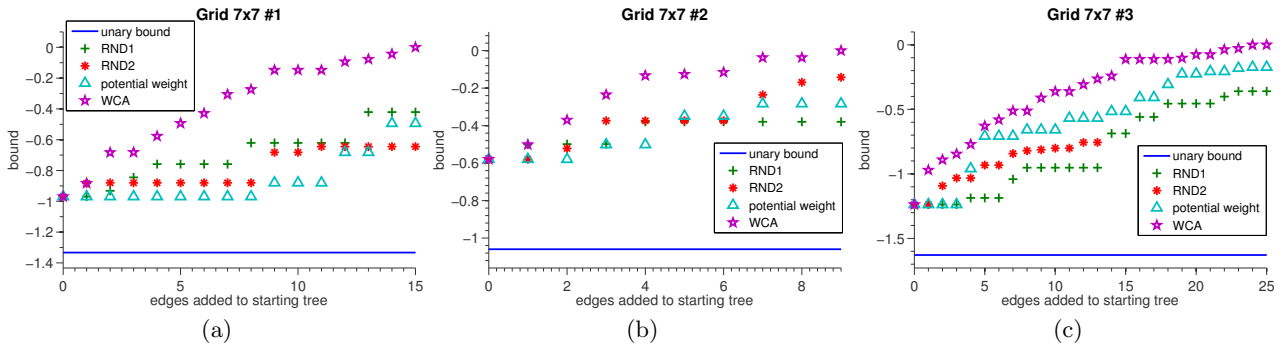

Figure 8.2: Sequentially adding edges. Comparison of the improvement in the dual bound after sequential edge addition using different criteria.

## 9 Discussion

Recent years have shown a resurgence of interest in higher-order potentials for graphical models with a growing number of specific potentials for which message-passing algorithms can be applied efficiently. In this paper we have shown that many of these methods are based on a particular linear programming relaxation and highlighted the weakness of that relaxation. We suggested a family of tighter relaxations which result in a practical new strategy that can yield significantly improved accuracy. The computational cost incurred for this increased accuracy is exponential in the tree-width of the consistency sets $S$, but empirically we see that substantial gains in accuracy can be achieved with relatively small tree-width.

One challenge for the cardinality HOP is scaling up to bigger problems. Regardless of the tree-width, there is a computational cost to the message computations that is quadratic in the number of variables in the model. When we apply our approach to large problems (e.g., images where individual pixels are variables), this cost becomes a bottleneck. A strategy that we would like to investigate is to use Fast Fourier Transforms near the zero-temperature limit to compute approximate max-marginals, or to investigate other algorithms for fast approximate max-convolution. Throughout, we assumed that the variables are binary and that the energy decomposes to unary and pairwise potentials with a single HOP. Extensions to the 3rd or 4th order cliques that are commonly used in computer vision would be straightforward by enforcing consistency between the HOP and all these cliques, and the assumption of binary variables can be removed so long as the HOP computations can be done tractably. For the case of multiple HOPs in the model, while there are open questions about some of the specifics (e.g., should different HOPs be constrained to choose the same sets of edges or not?), the same basic approach that we presented here is also applicable, and we believe it to be a good choice. Finally, we have shown how to efficiently compute message updates for two classes of HOPs. An open question and a new computational challenge is to discover other cases where messages can similarly be computed efficiently.

### Acknowledgements

This research is partially supported by the ISF Centers of Excellence grant 1789/11, CIfAR and the Charitable Gatsby Foundation.

## A  Proof that $\mathbf{LP}_\emptyset = \mathbf{LP_{red}}$

**PRIMAL $\mathbf{LP}_\emptyset \leq$ PRIMAL $\mathbf{LP_{red}}$**  Copy the solution from $\mathbf{LP_{red}}$ into $\mathbf{LP}_\emptyset$ for the variables that correspond, and set $q_\alpha(\mathbf{x}) = \sum_z \frac{\prod_{i'} q_{zi'}(z, x_{i'})}{q_z(z)^{n-1}}$, which is a distribution over $x$ and $z$ that has $q_{zi}(z, x_i)$ as its pairwise marginals, because this corresponds to the Bethe approximation on a tree-structured graph (in this case, a star around $z$), which is exact. Unary consistency in $\mathbf{LP}_\emptyset$ is satisfied because

$$\sum_{\mathbf{x}:\mathbf{x}(i)=x_i} q_\alpha(\mathbf{x}) = \sum_{\mathbf{x}:\mathbf{x}(i)=x_i} \sum_z \frac{\prod_{i'} q_{zi'}(z, x_{i'})}{q_z(z)^{n-1}} \quad \text{(A.1)}$$

$$= \sum_z q_{zi}(z, x_i) = q_i(x_i). \quad \text{(A.2)}$$

The objective of $\mathbf{LP}_\emptyset$ is less than or equal to the objective at $\mathbf{LP_{red}}$, because $\theta_\alpha(\mathbf{x}) = \min_z \sum_i \theta_{zi}(z, x_i)$:

$$\sum_\mathbf{x} q_\alpha(\mathbf{x})\theta_\alpha(\mathbf{x}) \quad \text{(A.3)}$$

$$= \sum_\mathbf{x}(\sum_z \frac{\prod_{i'} q_{zi'}(z, x_{i'})}{q_z(z)^{n-1}})(\min_z \sum_i \theta_{zi}(z, x_i)) \quad \text{(A.4)}$$

$$\leq \sum_\mathbf{x} \sum_z \frac{\prod_{i'} q_{zi'}(z, x_{i'})}{q_z(z)^{n-1}} \sum_i \theta_{zi}(z, x_i) \quad \text{(A.5)}$$

$$= \sum_i \sum_{z, x_i} q_{zi}(z, x_i)\theta_{zi}(z, x_i). \quad \text{(A.6)}$$

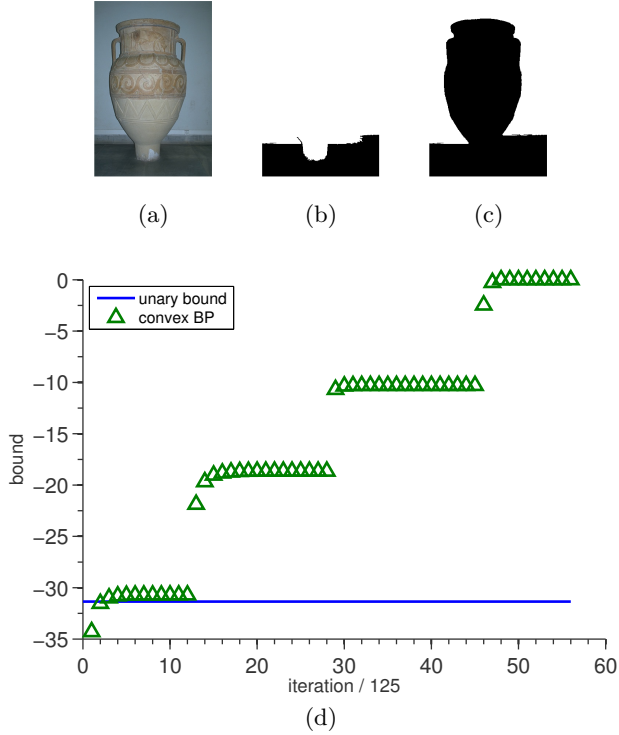

Figure 8.3: Image Segmentation example run. (a) Input image. (b) Suboptimal solution found by the commonly used spectral method. (c) The global optimum, which is found and verified by our algorithm. (d) The dual bound of convex BP versus iteration of message passing. Jumps occur when edges are added to $S$.

**DUAL $\text{LP}_\emptyset \geq$ DUAL $\text{LP}_{\text{red}}$** Here we take a dual solution from $\text{LP}_{\text{red}}$ and construct a dual objective for $\text{LP}_\emptyset$ that is greater than or equal to the $\text{DUAL}_{red}$ objective. For a given setting of dual variables $\delta$ and $\gamma$, the duals for $\text{LP}_\emptyset$ ($\text{DUAL}_\emptyset$) and $\text{LP}_{\text{red}}$ ($\text{DUAL}_{red}$), respectively, are as follows:

**$\text{DUAL}_{red}$**

$$\sum_i \min_{x_i}[\theta_i(x_i) + \sum_{j \in N(i)} \delta_{ji}(x_i) + \gamma_{zi}(x_i)] \quad (A.7)$$

$$+ \sum_{ij} \min_{x_i,x_j}[\theta_{ij}(x_i, x_j) - \delta_{ji}(x_i) - \delta_{ij}(x_j)]$$

$$+ \sum_i \min_{z,x_i}[\theta_{zi}(z, x_i) - \gamma_{zi}(x_i) - \gamma_{iz}(z)] + \min_z[\sum_i \gamma_{iz}(z)]$$

**$\text{DUAL}_\emptyset$**

$$\sum_i \min_{x_i}[\theta_i(x_i) + \sum_{j \in N(i)} \delta_{ji}(x_i) + \gamma_i(x_i)] \quad (A.8)$$

$$+ \sum_{ij} \min_{x_i,x_j}[\theta_{ij}(x_i, x_j) - \delta_{ji}(x_i) - \delta_{ij}(x_j)]$$

$$+ \min_{\mathbf{x}}[\theta_\alpha(\mathbf{x}) - \sum_i \gamma_i(x_i)]$$

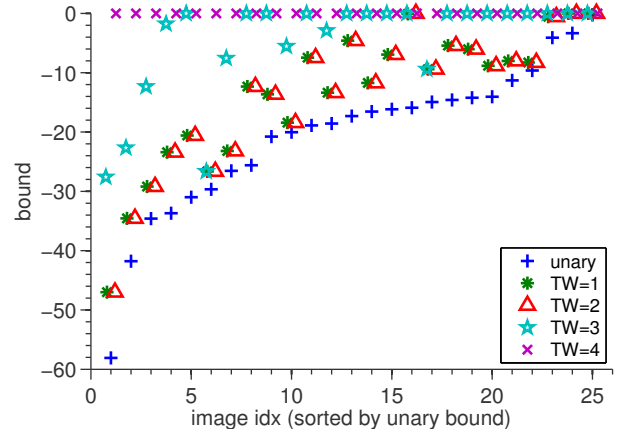

Figure 8.4: Lower bounds achieved by our algorithm versus the standard $\text{LP}_\emptyset$ bound ("unary") on average-cut image segmentation problems at various tree-widths of $S$, ranging from 1 to 4 ("TW-1" to "TW-4"). In all cases the optimal integral energy is 0, so when the lower bound reaches 0 we have provably reached the optimum.

Now copy the $\delta$ messages from $\text{DUAL}_{red}$ to $\text{DUAL}_\emptyset$. In $\text{DUAL}_\emptyset$, set $\gamma_i(x_i) = \gamma_{zi}(x_i)$. Then the difference in the dual objectives between Eq. A.8 and Eq. A.7 is only in the terms involving $z$ or $\alpha$. Focusing on the $z$ terms from Eq. A.7:

$$\sum_i \min_{z,x_i}[\theta_{zi}(z, x_i) - \gamma_{zi}(x_i) - \gamma_{iz}(z)] + \min_z[\sum_i \gamma_{iz}(z)] \quad (A.9)$$

$$\leq \min_{z,\mathbf{x}} \sum_i [\theta_{zi}(z, x_i) - \gamma_{zi}(x_i) - \gamma_{iz}(z)] + \min_z[\sum_i \gamma_{iz}(z)] \quad (A.10)$$

$$\leq \min_{z,\mathbf{x}} \sum_i [\theta_{zi}(z, x_i) - \gamma_{zi}(x_i)] \quad (A.11)$$

$$= \min_{\mathbf{x}}[\theta_\alpha(\mathbf{x}) - \sum_i \gamma_i(x_i)], \quad (A.12)$$

which shows that the dual objective for $\text{LP}_\emptyset$ is greater than or equal to the dual objective for $\text{LP}_{\text{red}}$ with this choice.

So we have $\text{DUAL}_\emptyset \geq \text{DUAL}_{red}$ and PRIMAL $\text{LP}_{\text{red}} \geq$ PRIMAL $\text{LP}_\emptyset$. Since strong duality for LPs gives $\text{DUAL}_\emptyset =$ PRIMAL $\text{LP}_\emptyset$ and $\text{DUAL}_{red} =$ PRIMAL $\text{LP}_{\text{red}}$, this implies that the two LPs have the same solution value and are thus equally tight.